\documentclass{article}
\usepackage{spconf,amsmath,graphicx}

\title{Blockwise Temporal-Spatial pathway network}
\name{SeulGi Hong and Min-Kook Choi}
\address{hutom, Seoul, South Korea}

\begin{document}
%
\maketitle
\begin{abstract}
Algorithms for video action recognition should consider not only spatial information but also temporal relations, which remains challenging. We propose a 3D-CNN-based action recognition model, called the blockwise temporal-spatial path-way network (BTSNet), which can adjust the temporal and spatial receptive fields by multiple pathways. We designed a novel model inspired by an adaptive kernel selection-based model, which is an architecture for effective feature encoding that adaptively chooses spatial receptive fields for image recognition. Expanding this approach to the temporal domain, our model extracts temporal and channel-wise attention and fuses information on various candidate operations. For evaluation, we tested our proposed model on UCF-101, HMDB-51, SVW, and Epic-Kitchen datasets and showed that it generalized well without pretraining. BTSNet also provides interpretable visualization based on spatiotemporal channel-wise attention. We confirm that the blockwise temporal-spatial pathway supports a better representation for 3D convolutional blocks based on this visualization.
\end{abstract}


\vspace{-2mm}
\section{Introduction}
\label{sec:intro}
This paper proposes a new video action recognition model that exploits a dynamic selection of receptive fields. When we widen the target domain to 3D, algorithms for video action recognition should consider both spatial and temporal information. In particular, temporal relations have been regarded as a key issue. Intuitively, every clip has several clusters of movement that may be deeply related to the semantics of a major action or only have minor contributions (such as movement of background or object locations, swinging arms, or finger movements). These movements often occur simultaneously in video clips. The compositions of simultaneous movements are interpreted as an action by the human vision. Therefore, every single timestep requires corresponding receptive fields to extract their own semantics. Therefore, \textbf{every single timestep} requires \textbf{corresponding receptive fields} to extract their own semantics.

\begin{figure}[t]
\begin{center}
   \caption{{\bf Visualization.} Normalized attention weights along the time axis. vertical axis notates the contributions of each pathway on specific time.}
   \label{fig:main}
\includegraphics[width=1.0\linewidth]{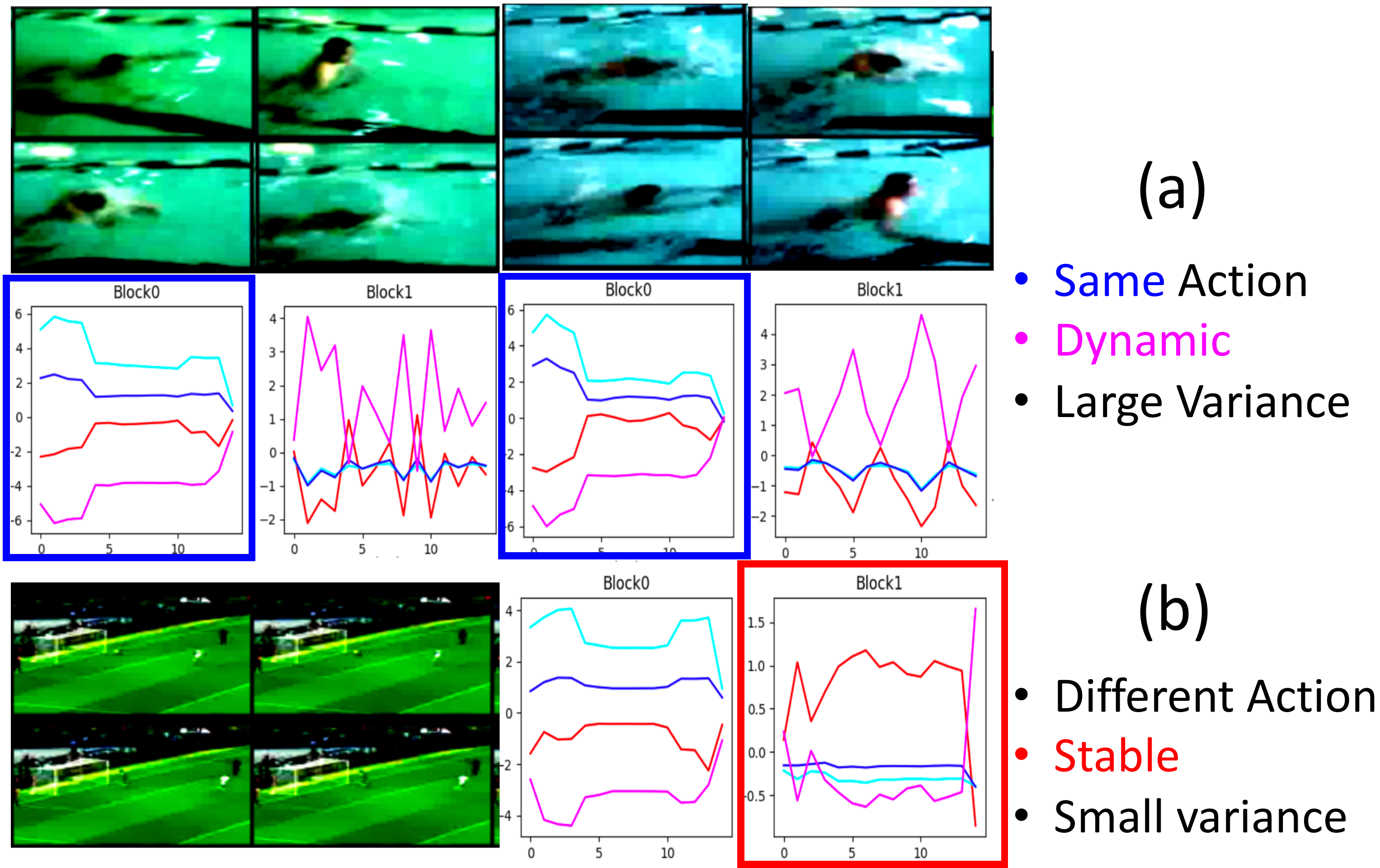}
\end{center}
\vspace{-7mm}
\end{figure}

  One of the previous methods SlowFast \cite{C2019slowfast}, can process entire clips at different frame rates by separating the model into two pathways, which have predefined receptive fields to cover both slow and fast motions. In other words, the receptive fields vary in two pathways with their own sampling rates. The feature pyramid \cite{Yang2020tpn} also attempts to capture temporal information. However, beyond extracting the features on multiple scales, the model must focus precisely on the informative temporal location.

In addition, backbone architectures are highly developed in the 2D domain, and one of the models shares the same aspect as previously mentioned. An existing study adaptively exploits receptive fields in an image domain \cite{li2019selective}. The importance of the receptive field in 2D images was emphasized in \cite{li2019selective}, and their proposed model learns which scale should be used. After the parallel convolution layers, it gives channel-wise attention weights to the features from square-shaped receptive fields extracted on multiple scales. Thus, feature maps are obtained from an informative spatial scale.

Inspired by the aforementioned studies, we extend these approaches and propose a new video action recognition model. Our suggested model, named blockwise temporal-spatial pathway network (BTSNet), dynamically allocates receptive fields for each timestep.
\begin{figure*}[t]
    \begin{center}
       \caption{\textbf{Left: Procedure of TSP block of BTSNet.} For any given feature map, we execute convolutions on various temporal-spatial receptive fields. A temporal-channel attention is applied as a fuse layer. \textbf{Right}: Corresponding receptive fields for the feature map. Pathways have top-1 contribution at each timestep.}
       \label{fig:pipeline}
    \includegraphics[width=1.0\linewidth]{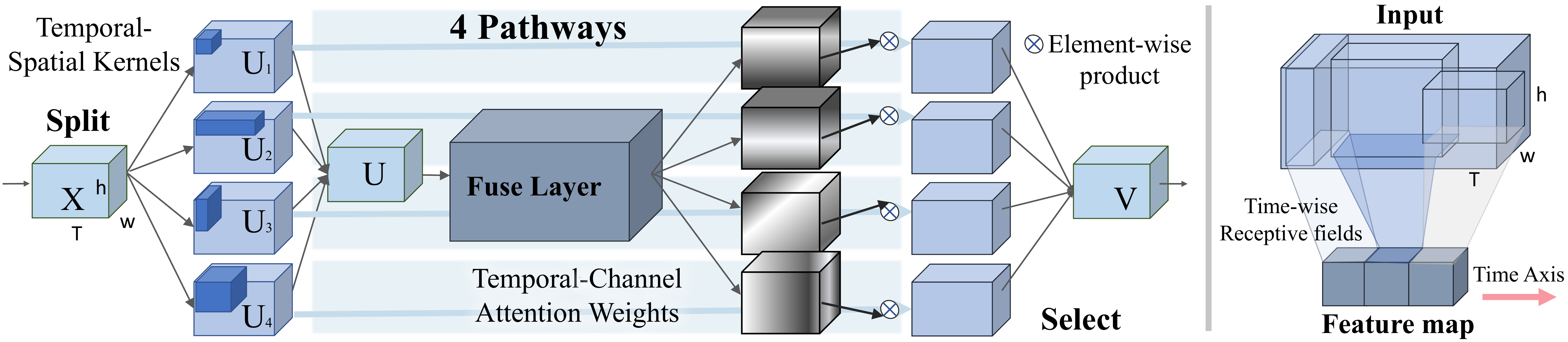}
    \end{center}
    \vspace{-9mm}
\end{figure*}
Although we share the concept of the pathway \cite{C2019slowfast} for processing the scale variances separately, our pathway is \textbf{a blockwise module}. The receptive fields of each convolutional layer are different because the upper layers have a larger scale. Therefore, our blockwise characteristics are important. Each block has a unique view on candidate receptive fields, as features are highlighted immediately after they are extracted from each spatiotemporal scale in a block-by-block manner. A fuse layer allows the proposed model to focus precisely on the informative receptive field (RF) per timestep in the blockwise module. These fused features are fed to the next block after weighting is performed. Note that the pathway in SlowFast \cite{C2019slowfast} is permanent because the receptive fields are predefined by sampling rates at the input. Through an ablation study, we determined that time-wise attention plays an important role, when both settings have channel-wise attention and exactly the same RF. The main contributions of our study are as follows:
\begin{itemize}
    \item{We propose a new video action recognition model. Specifically, we devise a blockwise pathway that can extract features from various temporal-spatial scales and process them effectively. In addition, our model is a generalized form that contains multiple pathways corresponding to receptive fields.}
    
    \item{Our model (BTSNet) provides interpretable visualization using the temporal-channel attention in our fuse layer to analyze the model’s focus. This visualization supports the evidence that the proposed model can achieve better generalization in 3D-CNN-based action recognition tasks.}
    
    \item{Our ablation experiments indicate that adaptive temporal receptive field is crucial. Although models have the same receptive fields with channel-wise attention, the results show that time-wise attention consistently enhances action recognition performance.
}
\end{itemize}

\vspace{-4mm}
\section{Related Work}
\vspace{-2mm}
\noindent \textbf{Spatiotemporal Receptive Fields.} Some existing models cover the temporal relation by expanding the convolution to 3D \cite{D2015c3d, Hara2017R3D, J2017quo, additional}. The kernels of 3D convolutions were decomposed in \cite{Tran2018ACL,Qiu2017PR3D}. TSN \cite{L2019tsn} uses a sparse temporal sampling strategy. In \cite{Feichtenhofer2016fusion}, the model fuses the temporal information by pooling. Instead of average pooling, \cite{Zhou2018trn} and \cite{Lin2019tsm} use a relational module and a shift module, respectively. We focus on temporal variance by multiple pathways, and SlowFast \cite{C2019slowfast} is intuitively similar to our concept. The main idea of SlowFast is to cover two different frame rates by defining slow and fast sampling rates for each pathway and fusing the information by a lateral connection. In our case, our generalized pathways can embrace multiple temporal-spatial scales using dilation parameters. TPN \cite{Yang2020tpn} also takes advantage of multiple scales that extract features from hierarchical layers. However, our model fuses the information blockwise to reflect which scale contributes more to each timestep.

\vspace{-3mm}
\section{Methods}
\begin{figure}[t]
\begin{center}
   \caption{{\bf Comparison of receptive fields of a convolution that corresponds to a pathway.} a) Pathway of SlowFast, which is manipulated by the sampling rate. b) Our suggested blockwise pathway, which is defined by dilation parameters.}
   \label{fig:RF}
\includegraphics[width=1.0\linewidth]{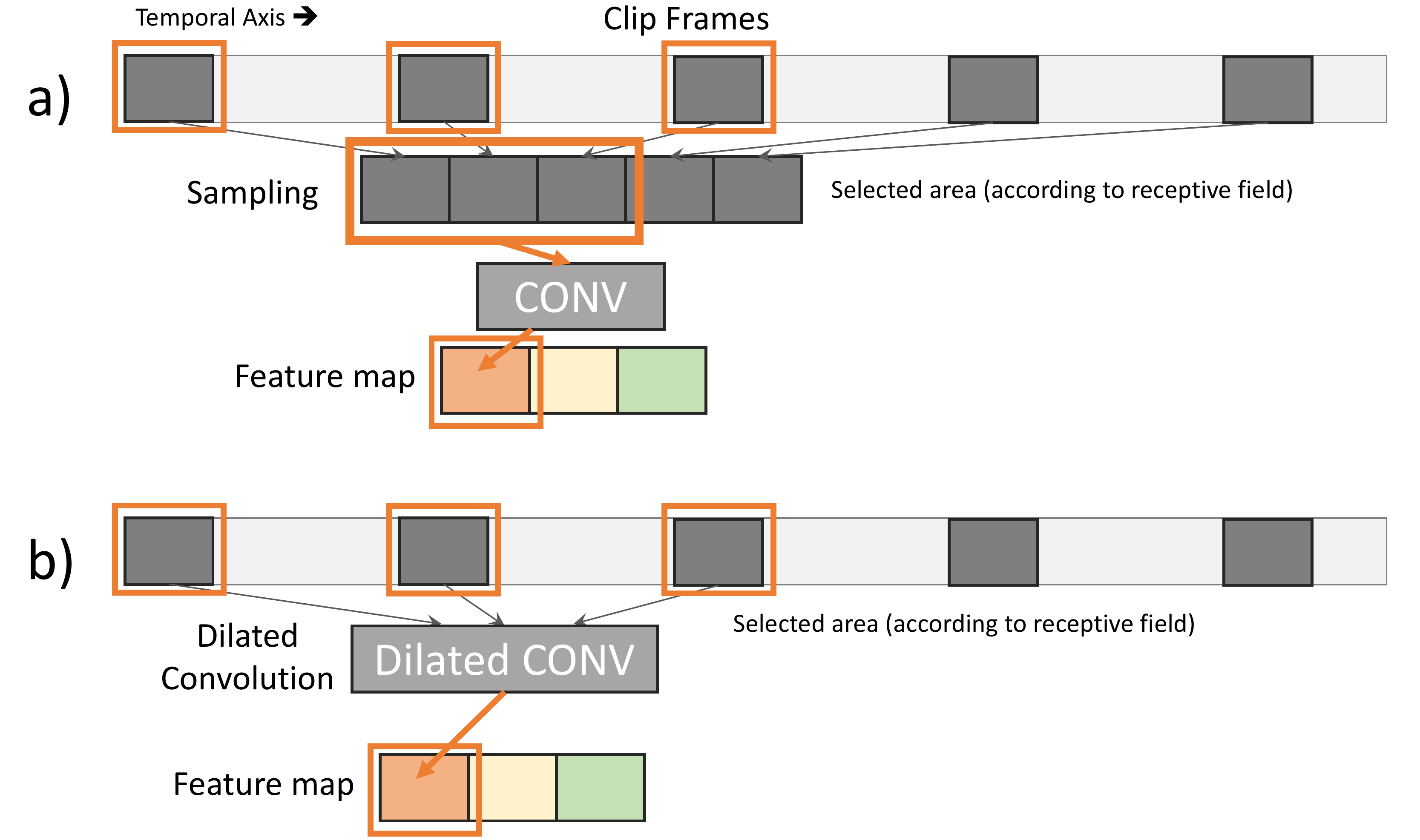}
\end{center}
\vspace{-7mm}
\end{figure}

\vspace{-3mm}
Our temporal-spatial pathway (TSP) block was inspired by \cite{li2019selective}. We expand it to the video domain and let the model consider temporal receptive fields. To this end, we defined various candidate operations with different receptive fields along the temporal and spatial axes. All procedures are described in Section 3.1 and Fig. \ref{fig:pipeline}. Moreover, we present the receptive fields of multiple pathways in Section 3.2 and Fig. \ref{fig:RF}.

\vspace{-2mm}
\subsection{Temporal-Spatial Pathway Block}
\vspace{-1mm}
The TSP block has 3 major operations: split, fuse, and select. This process is explained by dividing it into several steps.

\noindent \textbf{Split.} For any given feature map ${X\in{\rm I\!R^{C'\times T'\times H'\times W'}}}$, transformation functions $F_{1,2,..,m}$ are applied first. The number of functions is handled by a hyperparameter M. We use convolution as the function with a dilation parameter, which encourages every transformation to cover different receptive fields. Thus, our pathway blocks can be considered as slow, fast, or spatially enlarged pathways.
\begin{equation}
    F_m:X \to U_m\in{\rm I\!R^{C\times T\times H\times W}}
\end{equation}
\textbf{Fuse Layer.} There are two options for this layer: temporal-channel attention and channel-wise attention. The operations slightly differ according to this setting. To fuse information from multiple receptive fields, we combine the previous features by adding all $U_m$:
\begin{equation}
    U=\sum_{m=1}^M U_m
\end{equation}
Global average pooling (GAP) is applied to compress features U. The range of this operation differs according to the type of this step. We apply (3) when we use temporal-channel attention as a fuse layer or (4) for channel-wise attention.
\begin{equation}
    S_{c,t}=F_{gap}(U) = \frac{1}{H \times W}\sum_{j=1}^H\sum_{i=1}^W U(i,j)
\end{equation}
\begin{equation}
    S_{c}=F_{gap}(U) = \frac{1}{T\times H \times W}\sum_{t=1}^T\sum_{j=1}^H\sum_{i=1}^W U(i,j,t)
\end{equation}
Then, a compact feature ${Z\in{\rm I\!R^{d\times T}}}$ or ${Z\in{\rm I\!R^d}}$ can be attained by a set of operations $F_o$. Moreover, $F_{c1}$ is a convolution with a $(1,1,1)$ kernel for compressing a dimension, $BN$ is a batch normalization, and $sigma$ is a ReLU activation function. We follow the rule of determining d from \cite{li2019selective}.
\begin{equation}
    Z=F_{o}(S)=\sigma(BN(F_{c1}(S)))
\end{equation}
Z should be resized to ${Z'\in{\rm I\!R^{M\times C}}}$ or ${Z'\in{\rm I\!R^{M\times C\times T}}}$ to attain the attention vectors, by applying convolution $F_{c2}$ with kernel size 1.

\noindent \textbf{Select.} To highlight the information among multiple pathways, we applied softmax to adopt the attention mechanism. The temporal-channel or channel-wise attention weights were attained in this procedure.
\begin{equation}
    Attn=softmax(Z')
\end{equation}
Then, the attention vectors were split along M dimensions. These attention weights emphasize each pathway along the temporal-channel axis, which has a different receptive field. The final output V of the block is:
\begin{equation}
    V=\sum_{m=1}^M Attn_m*U_m.
\end{equation}

\subsection{Receptive Fields}
The schematic visualization in Fig. 3 indicates that our blockwise convolution can cover the area that SlowFast \cite{C2019slowfast} sees. Both the SlowFast and TSP blocks have a widened view along the temporal axis. However, the TSP block has more generalized RFs and can control the contributions of the RFs at each timestep. Two hyper-parameters to handle the RF in our pathway block: the number of pathways M and the RF option. For the RF option, we tried two different types of candidate operations to manipulate the receptive fields. In option 1 (O1), each path-way has a cube-like RF, increasing the temporal and spatial RFs proportionally. The dilation parameters $(T, H, W)$ are defined as follows:
\begin{equation}
    D=\{D_1, D_2,\cdots, D_M\}, D_i=(i,i,i).
\end{equation}
In option 2 (O2), each pathway has its own role, which means it can acquire visual information aided by various views to catch multiscale motions on the temporal or spatial axis. Various dilation parameters were defined manually. For instance, we set $\{{(1,1,1), (4,4,4), (1,4,4), (4,1,1)}\}$ when M is 4.


\vspace{-2mm}

\section{Experiments}
\vspace{-2mm}
\textbf{Settings.} We use the ResNeXt-based block for our structure, and 3D convolution is replaced with the proposed TSP block. The numbers of blocks for BTSNet-26, 50, and 101 are set as [2,2,2,2], [3,4,6,3], and [3,4,23,3], respectively. The number of parameters is listed in Table \ref{tab:params}. We do not use the TSP block for the last layer in Table \ref{tab:exp_report}.
To measure generalization capability, we evaluated our model without any pretraining. For the video action recognition task, we chose UCF-101 \cite{UCF}, HMDB-51 \cite{HMDB} and SVW \cite{SVW} datasets. In addition, we used Epic-Kitchen \cite{epic} to classify nouns and verbs separately, including 26,074 and 2,398 action segments for training and validation, respectively. We set baselines as SlowFast \cite{C2019slowfast} and 3D-ResNet \cite{Hara2017R3D} because 3D-ResNet is a representative model, and the concept of our model is deeply related to SlowFast. For a fair comparison, we experimented on a fixed setting, but the SGDR scheduler \cite{SGDR} is only used for SlowFast in UCF-101 and SVW because of the model convergence problem. Please refer to supplementary for setting details.

\begin{table}[t!]
  \caption{\textbf{Number of trainable parameters (millions).} Cardinality of our models is included in notation: C16 and C32.}
  \label{tab:params}
  \resizebox{\linewidth}{!}{
  \begin{tabular}{c|c|c|c|c|c|c}
\hline
   Model & R3D-50 & R3D-101 & RX32-50 & RX32-101 & - & - \\ 
   \hline
    \# & 46.4 & 85.5  & 26.1 & 47.7 & - & - \\
    \hline
    Model &  SF-50 & SF-101 & SF-152 & SF-200 & - & - \\
    \hline
    \# & 33.8 & 62.1 & 85.0 & 89.5 & - & - \\
   \hline	\hline
	Model & C16-26 & C16-50 & C16-101 & C32-26 & C32-50 & C32-101 \\
	\hline
    \# & 10.2 & 17.4 & 34.6 &  17.3 & 31.7 & 66.1 \\
	\hline
	\end{tabular}
}
	\vspace{-4mm}
\end{table}

\begin{table}[t!]
  \caption{\textbf{Validation accuracy on various datasets.} Top-1 accuracy ($\%$) for each dataset and architecture.}
  \label{tab:exp_report}
  \resizebox{\linewidth}{!}{
  
  \begin{tabular}{c|c|c|c|c|c}
\hline
   Dataset & UCF-101 & HMDB-51 & SVW & \multicolumn{2}{c}{Epic-Kitchen} \\ 
   \cline{5-6}
    & & & & noun & verb \\
   \hline \hline
	R3D-50 & 52.921 & 20.420 & 65.406 & 24.729 & 40.847 \\
	R3D-101 & 54.917 & 18.976 & 63.042 & 23.791 & 39.637 \\
	\hline
	X3D-50-C32 & 57.864 & 20.519 & 65.406 & 25.521 & 40.263 \\
	X3D-101-C32 & 56.397 & 23.211 & 63.042 & 23.499 & 39.658 \\
	\hline
	SF-50 & 53.199 & 24.557 & 68.519 & 22.686 & 38.157 \\
	SF-101 & 51.176 & 21.307 & 65.091 & 22.373 & 37.406 \\
	SF-152 & 52.181 & 22.062 & 66.824 & 21.706 & 39.408 \\
	SF-200 & 56.582 & 22.882 & 63.002 & 22.894 & 39.450 \\
    \hline
    \hline
	BTS-26-C16 & \textbf{58.829} & 24.327 & \textbf{69.582} & 25.083 & 40.450 \\
	BTS-50-C16 & 55.710 & 23.047 & 69.149 & 24.187 & 40.596 \\
	BTS-101-C16 & 56.979 & 21.044 & 64.460 & 24.812 & 38.157 \\
    \hline
	BTS-26-C32 & 58.684 & \textbf{24.918} & 69.307 & 25.751 & \textbf{41.222} \\
	BTS-50-C32 & 58.023 & 22.357 & 68.361 &\textbf{ 26.168} & 41.159  \\
	\hline
    \end{tabular}
}
\vspace{-7mm}
\end{table}

\subsection{Action Recognition Performance}
 We report the validation accuracy of the final epoch on split 1 of each action recognition dataset. For our model, we use M = 4, RF option 2, temporal-channel attention for fuse layer (TC), and ResNeXt cardinality of 32. The experimental results for UCF-101, HMDB-51, SVW, and EpicKitchen are shown in Table 2. The results indicate that our model outperforms the existing models.
 
Table \ref{tab:ab_M} shows the results of an ablation study on three elements: the number of blockwise pathways M, type of the fuse layer, and RF option. We tested on UCF-101 split 1 and fixed the ResNeXt cardinality size as 32. The top part of Table 3 indicates that temporal channel attention is crucial for obtaining informative features. Although we have enough pathways (M that is larger than 2), the fuse layer with temporal-attention (TC) works better on both of our receptive field options.

The middle left part of Table 3 shows that a sufficient number of pathways, M, is necessary: there is a large performance gap between M = 2 and above. This tendency also exists under other settings (the middle right part). To summarize, a setting with three or four pathways seems sufficient to cover the view of our model.

The bottom part of Table 3 shows the results for RF option settings. The results indicate that our receptive fields with various aspects (O2) perform better in specific settings. Despite no significant difference when using temporal-channel attention in the fuse layer (last two columns), the model training would fail when there is no time-wise attention with cube-shaped RFs (O1). To summarize, our model seems to attain informative areas when there are enough pathways with temporal-channel attention, even when there is no any spatial- or temporal-only receptive fields.

\begin{table}[t!]
  \caption{\textbf{Ablation study.} \textbf{Row 1.} fuse layer. \textbf{Row 2.} the number of pathways M. \textbf{Row 3.} RF options.}
  \label{tab:ab_M}
    \resizebox{\linewidth}{!}{
  \begin{tabular}{|c|c|c||c|c|c|}
    \hline
   Ablation & M3-O2-26 & M4-O2-26 & M3-O1-26 & M3-O1-50 & M3-O1-101\\
   \hline
	 C & 57.507 & 59.913 & 58.974 & 58.459 & 58.565 \\
	TC & 60.283 & 60.058 & 61.829 & 59.120 & 60.005 \\
	\hline
	TC-C & \textbf{2.776} & \textbf{0.145} & \textbf{2.855} & \textbf{0.661} & \textbf{1.440} \\
	\hline
    \end{tabular}}
    \resizebox{\linewidth}{!}{
  \begin{tabular}{|c|c|}
    \hline
   Ablation & TC-O2-50 \\
   \hline
	 M=2 & 55.855 \\
	M=3 & 58.895 \\
	M=4 & 58.274 \\
	M=7 & 58.551 \\
	\hline
	Max Gap & 3.04 \\
	\hline
    \end{tabular}
  \begin{tabular}{|c|c|c|c|}
    \hline
   Ablation & TC-O1-26 & TC-O1-50 & TC-O1-101\\
   \hline
	 M=2 & 60.019 & 58.842 & 56.133 \\
	M=3 & \textbf{61.829} & \textbf{59.120} & \textbf{60.005} \\
	\hline
   Ablation & C-O2-26 & C-O2-50 & C-O1-26\\
   \hline
	 M=3 & 57.507 & 57.454 & 58.974 \\
	M=4 & \textbf{59.913} & \textbf{58.261} & \textbf{59.252} \\
	\hline
    \end{tabular}
}
  \resizebox{\linewidth}{!}{
  \begin{tabular}{|c|c|c|c|c|}
    \hline
   Ablation & C-M4-26 & C-M4-50 & TC-M3-50 & TC-M4-50\\
   \hline
	 O1 & 59.252 & 48.678 & 59.120 & 58.961 \\
	O2 & 59.913 & 58.261 & 58.895 & 58.274 \\
	\hline
	O2-O1 & \textbf{0.661} & \textbf{9.583} & -0.225 & -0.687 \\
	\hline
    \end{tabular}
}
\vspace{-4mm}
\end{table}

We also experimented with model depth. However, it is difficult to obtain any benefits from the model depth. The optimal model depth depends on the datasets or tasks.

\begin{figure}[t]
\begin{center}
   \caption{{\bf Visualization.} Normalized attention weights along the time axis.}
   \label{fig:vis}
\includegraphics[width=1.0\linewidth]{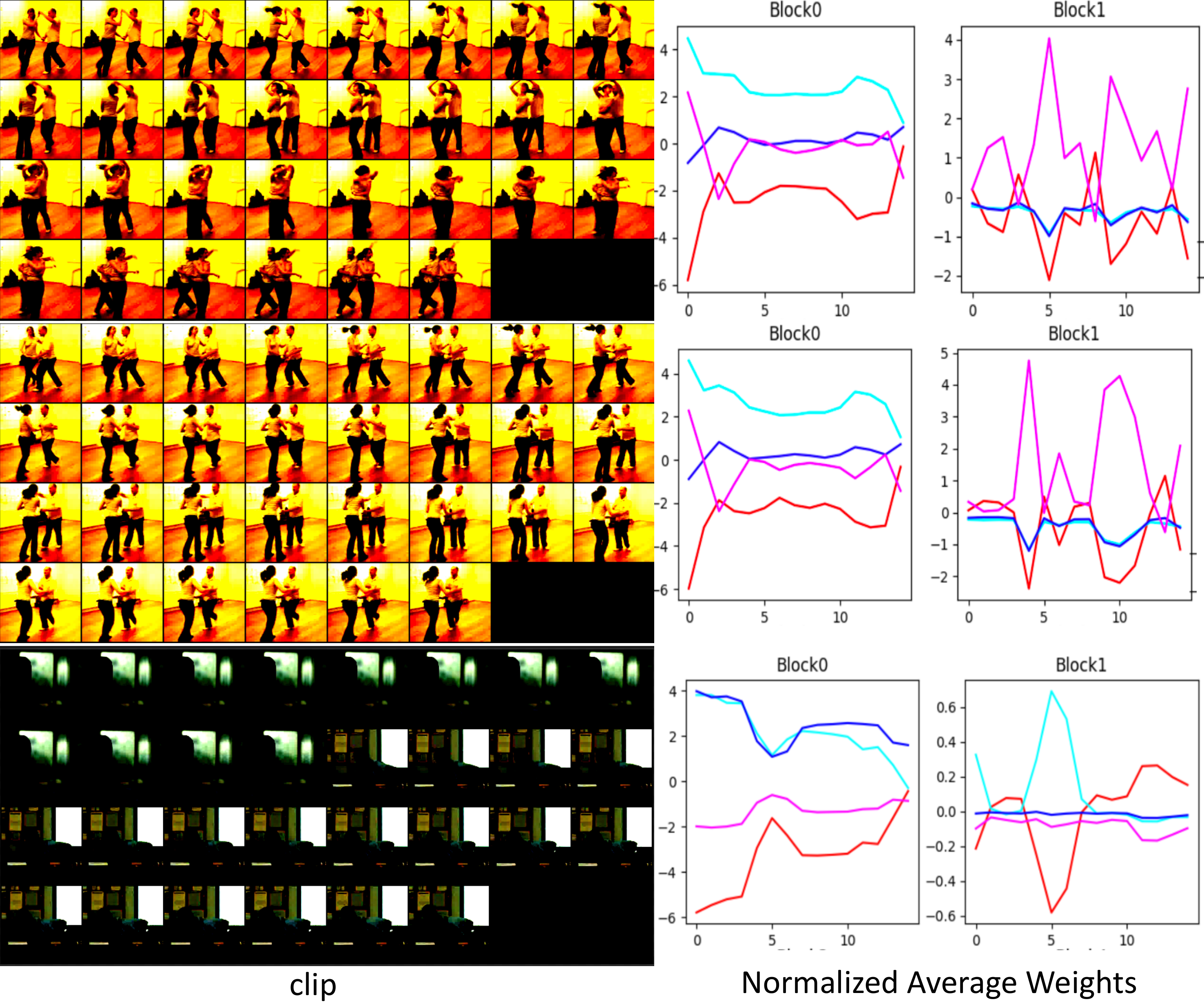}
\end{center}
\vspace{-10mm}
\end{figure}

\subsection{Visualization}
We visualized the temporal-channel attention weights for the validation set. First, attention weights are allocated well to the corresponding timesteps of each clip and are not biased. Second, the first layer (block0) seems to prefer spatially larger RFs, and the next layer (block1) tends to consider the variance of frame rates. Third, the distributions of weights have intra-class similarity. Furthermore, the inferenced weights have a large variance when the clip has dynamic actions. To summarize, we confirm that temporal-spatial channel-wise attention helps to effectively represent compounded action in video clips depending on each convolutional block of the 3D-CNN-based model. For more examples, please refer to the supplementary.

\vspace{-4mm}

\section{Conclusion}
We proposed BTSNet, a novel architecture for effective action recognition, which can adaptively manage the contribution of receptive fields for each timestep in a video clip. The experiments show that our time-wise receptive field approach is crucial, and visualization also indicates that the importance of receptive fields is considerably different over time.

\vspace{\baselineskip}

\noindent \textbf{Acknowledgements:} This work was supported by the Korea Medical Device Development Fund grant funded by the Korea government (the Ministry of Science and ICT, the Ministry of Trade, Industry and Energy, the Ministry of Health \& Welfare, the Ministry of Food and Drug Safety) (Project Number: 202012A02-02)


\bibliographystyle{IEEEbib}
\bibliography{strings,refs}

\end{document}